\documentclass[10pt,twocolumn,letterpaper]{article}

\usepackage[final]{cvpr}              
\usepackage{booktabs}
\usepackage{svg}
\svgsetup{inkscapelatex=false}

\definecolor{cvprblue}{rgb}{0.21,0.49,0.74}
\usepackage[pagebackref,breaklinks,colorlinks,citecolor=cvprblue]{hyperref}

\def\paperID{1253} 
\def\confName{CVPR}
\def\confYear{2024}

\title{Principal Component Clustering for Semantic
Segmentation in Synthetic Data Generation}

\author{Felix Stillger\\
Bergische Universität Wuppertal, Aptiv\\
Wuppertal, Germany\\
{\tt\small felix.stillger@uni-wuppertal.de}
\and
Frederik Hasecke\\
Aptiv\\
Wuppertal, Germany\\
{\tt\small frederik.hasecke@aptiv.com}
\and
Tobias Meisen\\
Bergische Universität Wuppertal\\
Wuppertal, Germany\\
{\tt\small meisen@uni-wuppertal.de}
}

\begin{document}
\maketitle
\section{Introduction}
\label{sec:intro}
This technical report outlines our method for generating a synthetic dataset for semantic segmentation using a latent diffusion model, as described in  \cite{rombach2022highresolution}. Our approach eliminates the need for additional models specifically trained on segmentation data and is part of our submission to the CVPR 2024 workshop challenge, entitled \textit{CVPR} 
2024 workshop challenge \textit{"SyntaGen - Harnessing Generative Models for 
Synthetic Visual Datasets"}. In the following, we describe the development of our pipeline and the training of a \textit{DeepLabv3} \cite{chen2017rethinking} model on the generated dataset. 
Our methodology is inspired by the attention interpretation techniques used in \textit{Dataset Diffusion} \cite{nguyen2023dataset}. To implement the 
self-attention gathering, we leverage the \textit{DAAM} \cite{tang2022daam} block. These self-attentions facilitate a novel head-wise semantic information condensation, thereby enabling the direct acquisition of class-agnostic image segmentation from the \textit{Stable Diffusion latents}.
Furthermore, we employ \textit{OVAM} \cite{marcosmanchon2024openvocabulary} for 
non-prompt-influencing cross-attentions from text to pixel, thus facilitating the classification of the previously generated masks. Finally, we propose a mask refinement step by using only the output image by \textit{Stable Diffusion}.
Our code is available on \textit{GitHub}\footnote[1]{\url{https://github.com/felixstillger/Syntagen_Submission_PCC_Segmentation}}.

\section{Method}
\label{sec:method}

\begin{figure}[t]
    \centerline{\includesvg[width=8cm]{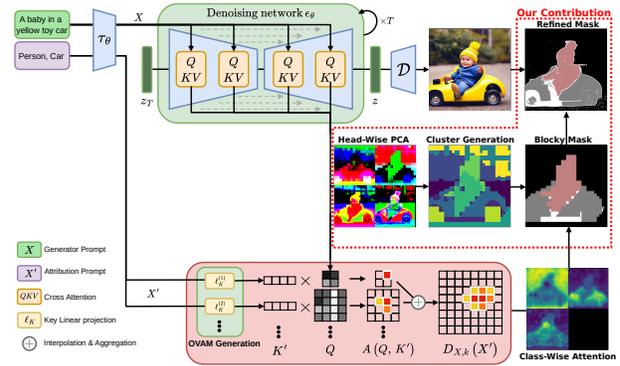}}
    \caption{Diagram of the Pipeline (Figure Adapted from \textit{OVAM} \cite{marcosmanchon2024openvocabulary})}
    \label{fig_diagram} 
  \end{figure}

Our methodology employs \textit{Stable Diffusion 2.1} \cite{rombach2022highresolution} as the foundational model, leveraging publicly available text prompts from \textit{Dataset Diffusion} \cite{nguyen2023dataset}, which are originally derived from the \textit{Pascal VOC} \cite{pascal-voc-2012} training dataset. Unlike \cite{nguyen2023dataset} and similar approaches that typically average and sum self-attentions across all heads and timesteps to manage these large tensors, our approach considers each head's features independently in the final iteration step. This allows us to generate masks that isolate the semantic content of the output image. The overall pipeline is illustrated in \textsc{Figure} \ref{fig_diagram}, with our contribution highlighted in the red dotted area.

\subsection{Process Self-Attentions}
Each head in the denoising network targets distinct features and objectives. We hypothesize that the attention mechanism, being critical for image reconstruction, conveys semantic information. Hence, our objective is to extract semantic meanings that are interpretable by humans and correspond specifically to the \textit{Pascal VOC} classes. 
To achieve this, we employ \textit{Principal Component Analysis} (\textit{PCA}) for each head, reducing 
the feature dimensionality from 64 to 3. This procedure is repeated at each upsampling layer, from a resolution of $16\times16$ up to $64\times64$. By performing such head-specific PCA, we condense the original features, facilitating the separation of the most distinct objectives for each head.

\subsection{Clustering and Classification}
After computing the principal components for each head, we upsample all smaller outputs to a resolution of $64\times64$. Subsequently, we concatenate the principal components of each head for every individual pixel to form a single tensor. To enhance robustness, we normalize the features and include the relative pixel positions as additional features. As, according to the challenge rules, the use of annotated masks is prohibited, we employ unsupervised \textit{K-Means clustering} multiple times with varying numbers of clusters. In doing so, this approach generates clusters by minimizing the squared Euclidean distance of the head-wise principal components. To ensure reproducibility, we fix the random initialization of clusters.

Determining the optimal number of clusters is complex as it depends on multiple factors, which cannot be exhaustively defined and extend beyond class, scene, and semantic meaning. To address this and maintain flexibility, we cluster using [$4, 7, 10$] clusters, aiming for a rough segmentation of main objects and a finer segmentation of smaller parts. We separate clusters that are not directly connected via 4-connected neighborhood pixels.

For class assignment to the masks, we utilize cross-attention maps from \textit{OVAM} \cite{marcosmanchon2024openvocabulary}, enabling class identification independent of the input prompt. We avoid optimized tokens to keep the approach generalized and free from any segmentation labels. Additionally, we observed that the \textit{start-of-text} (\textit{SoT}) token can serve as an indicator for the background class. Furthermore, we replace certain class names with more descriptive token names (see \textsc{Table} \ref{tab:renames}).

\begin{table}[htbp]
    \caption{Renaming Class Names to More Expressive Tokens.}
    \label{tab:renames}
    \begin{tabular}{c|c}
    \toprule
    \textbf{Pascal VOC Name} & \textbf{Renamed for OVAM} \\
    \midrule
    diningtable& table \\
    tvmonitor& monitor\\
    pottedplant& pot plant\\
    aeroplane& airplane\\
    \bottomrule
    \end{tabular}
    \centering
  \end{table}
  
We iterate over all clusters and apply varying confidence thresholds to the class-wise cross-attention maps from \textit{OVAM}. Initially, we normalize all original attention values to the range [0, 1], followed by multiple binary thresholding at 0.3, 0.5, and 0.8. For the background class, derived from the \textit{SoT} token, we increase the confidence threshold by 20\% of the current threshold to mitigate the excessive influence of the background. For the final classification, we compute the Intersection over Union (\textit{IoU}) for each cluster against the class-wise binary map, assigning the class with the highest \textit{IoU} to all pixels within that cluster for the defined thresholds. This process is repeated across all clusters and thresholds, with the most frequent class for each pixel determined by the mode. If the dominant class for a pixel constitutes 50\% or less, the pixel is labeled as uncertain. This procedure yields a $64\times64$ resolution mask. To enhance this low-resolution mask to the image resolution, we utilize the output image knowledge from \textit{Stable Diffusion}.

\subsection{Mask Refinement}

All previous operations were exclusively applied to the extracted latents of Stable Diffusion. For the final mask refinement, we utilize the RGB values of the output image and reintroduce the pixel positions. We employ \textit{K-Means clustering} with a fixed cluster count of 20 and separate disconnected clusters. This approach allows us to identify regions that are coherent in terms of color and proximity, irrespective of the image's semantic content. A class is assigned to these new clusters if the dominant class constitutes more than 66\% of the pixels within the cluster, ensuring accurate classification with high confidence. Due to the large and fixed cluster count, the \textit{K-Means clustering} algorithm may generate artifacts in certain segmentation masks. These artifacts are filled with the uncertainty class, and such pixels are excluded from further influence during training.
\section{Results}
\label{sec:results}
To evaluate the effectiveness of our dataset generation method, we produced just under $10 ,\!  000$ images for our final submission to the challenge, constrained by time limitations. To assess the quality of the synthetic dataset, we exclusively trained a \textit{DeepLabv3} model \cite{chen2017rethinking} on the generated images. We identified the best-performing checkpoint by calculating the \textit{mean Intersection over Union} (\textit{mIoU}) on the \textit{Pascal VOC} validation set \cite{pascal-voc-2012} at every 1000 training iterations. The checkpoint exhibiting the highest \textit{mIoU} was submitted to the challenge organizers, accompanied by its checksum.

\begin{table}[ht]
    \centering
  	\caption{Submission Results}
    \label{tab:sub}
    \begin{tabular}{c|c}
        \toprule
        \textbf{Submission}& \textbf{mIoU} \\
		\midrule
        $\dagger$ Teddy Bear (HCMUS)&51.66 \\
        HNU-VPAI  & 47.36 \\
        \textbf{Hot Coffee (Ours)} & \textbf{46.25} \\
        CVTEAMQ & 45.88 \\
        \midrule
        Dataset Diffusion (Baseline)& 46.85    \\
        \bottomrule
    \end{tabular} \\
    \centering
   	\begin{minipage}{0.3\textwidth}\footnotesize
   	\vspace{5pt}
   	$\dagger$ The entry was disqualified due to accidental use of segmentation labels in a module. The listed score is the postmortem correction.
   \end{minipage}
  \end{table}

In \textsc{Table} \ref{tab:sub} the results on the private test set are presented. Our submitted model performed slightly below the baseline established by Nguyen et al.\cite{nguyen2023dataset}, with a  difference of $0.4$ \textit{mIoU}. Some illustrative examples are provided in the appendix in \textsc{Figure} \ref{fig:results}.
Further, a comprehensive analysis of the performance of a \textit{DeepLabv3} model trained on our dataset is detailed in \textsc{Table} \ref{tab:comparison} in the appendix, encompassing both the \textit{private Syntagen} and \textit{Pascal VOC} validation set.

\section{Conclusion}
Certain classes in the dataset exhibit varying levels of difficulty. Our method encountered significant challenges with the \textit{person} 
class due to its close embedding with other classes, which resulted in a lower \textit{IoU}. Structurally challenging classes, such as sofa and \textit{chair}, also presented difficulties. 
In \textit{Pascal VOC}, the primary distinction between these two classes is that sofas are typically two-seaters, while chairs are usually 
one-seaters. This subtle difference seems to be challenging for the model to accurately discern.
These challenges underscore the need for enhanced prompt engineering to improve performance. 

Overall, we have developed a novel approach for semantically segmenting images generated by Stable Diffusion by leveraging only their latent representations. Our method is particularly effective for classes such as \textit{potted plant, bird, dog, cat, sheep, boat}, producing highly precise masks  (see \textsc{Figure} \ref{fig:results} in the appendix). This precision is achieved by utilizing head-wise features from both the latent and overall image spaces.

{
    \small
    \bibliographystyle{ieeenat_fullname}
    \bibliography{main}
}
\section*{Appendix}

\begin{figure}[ht]
  \begin{center}
    
  {\includegraphics[width=1.6cm, height=1.6cm]{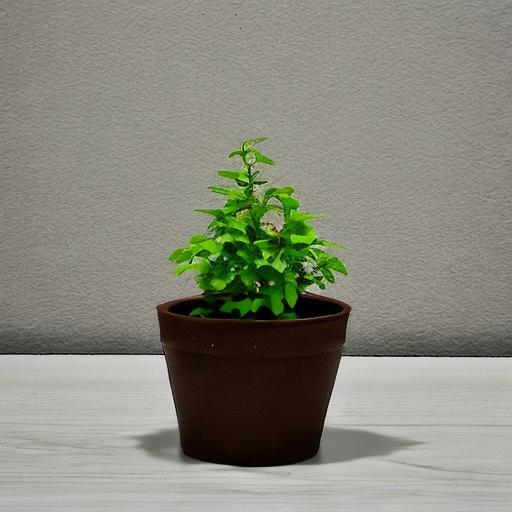}}
  {\includegraphics[width=1.6cm, height=1.6cm]{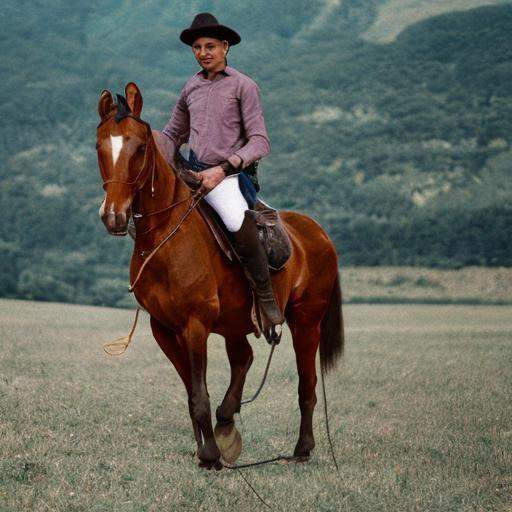}}
  {\includegraphics[width=1.6cm, height=1.6cm]{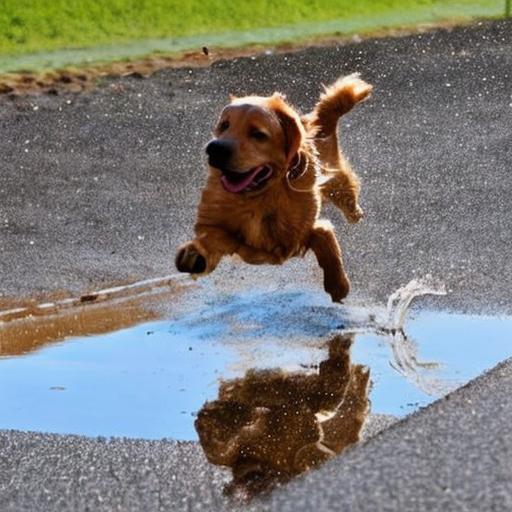}}
  {\includegraphics[width=1.6cm, height=1.6cm]{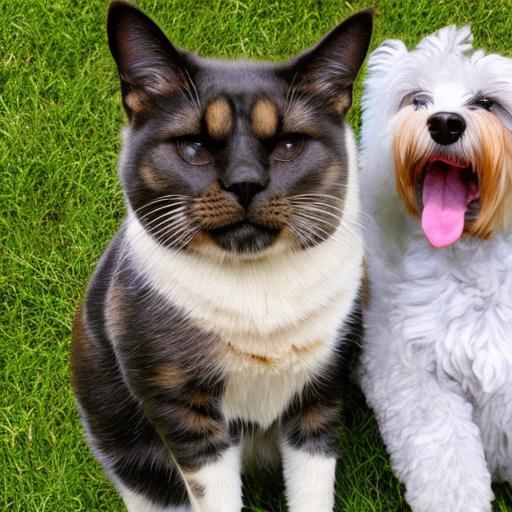}}
  {\includegraphics[width=1.6cm, height=1.6cm]{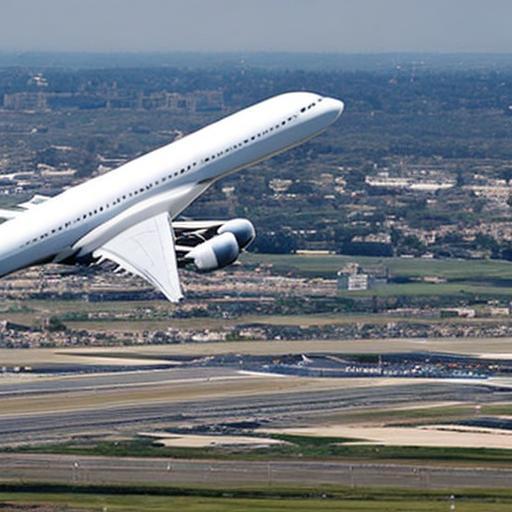}}

    {\includegraphics[width=1.6cm, height=1.6cm]{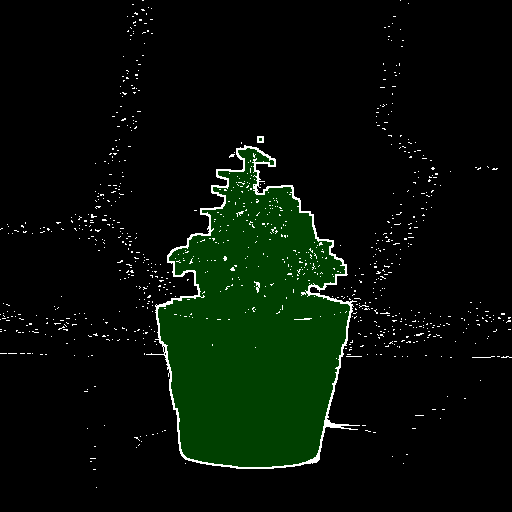}}
    {\includegraphics[width=1.6cm, height=1.6cm]{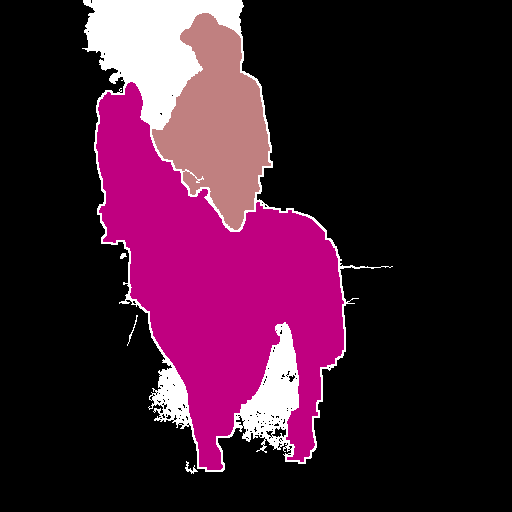}}
    {\includegraphics[width=1.6cm, height=1.6cm]{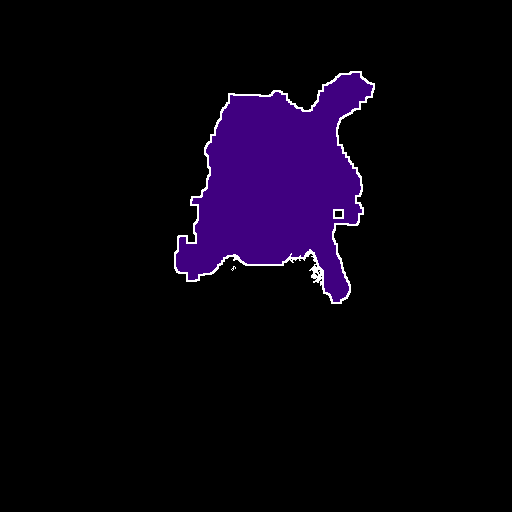}}
    {\includegraphics[width=1.6cm, height=1.6cm]{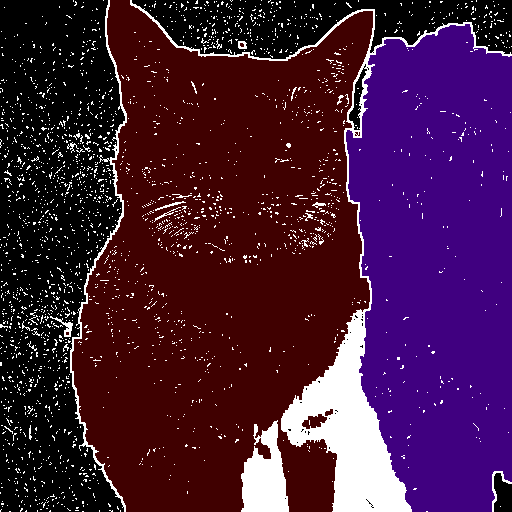}}
    {\includegraphics[width=1.6cm, height=1.6cm]{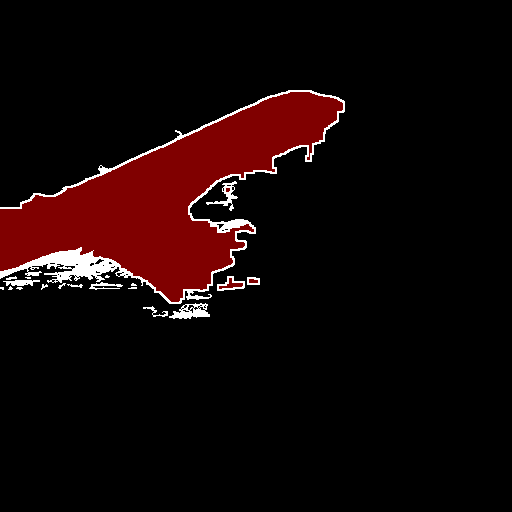}}
  \end{center}
  \begin{center}

    {\includegraphics[width=1.6cm, height=1.6cm]{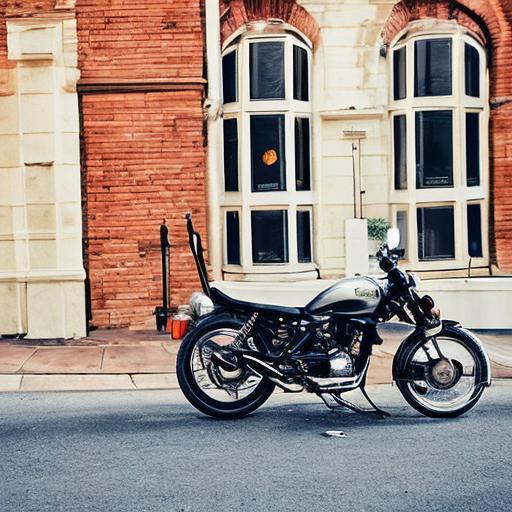}}
    {\includegraphics[width=1.6cm, height=1.6cm]{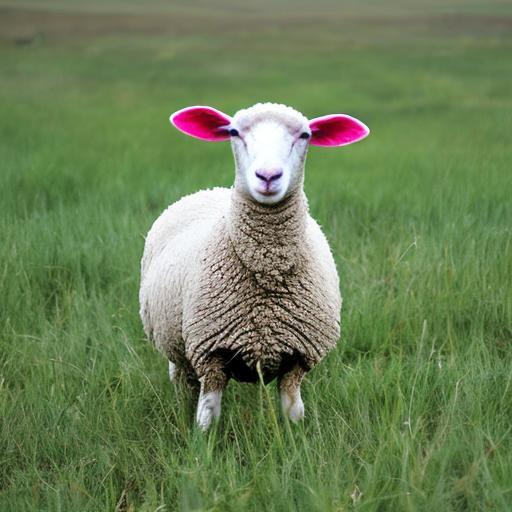}}
    {\includegraphics[width=1.6cm, height=1.6cm]{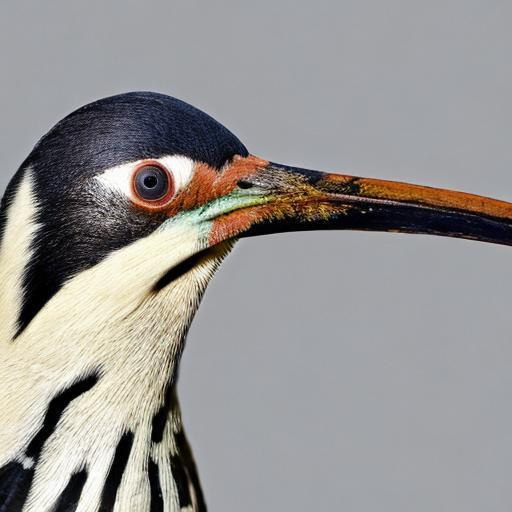}}
    {\includegraphics[width=1.6cm, height=1.6cm]{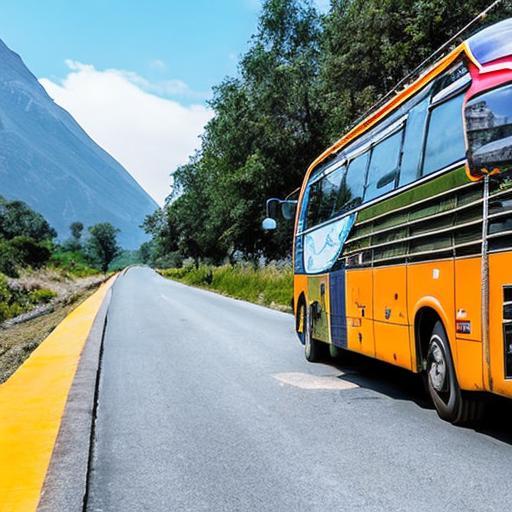}}
    {\includegraphics[width=1.6cm, height=1.6cm]{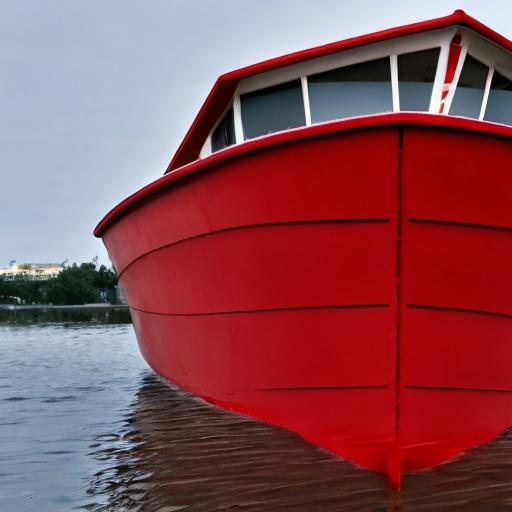}}

      {\includegraphics[width=1.6cm, height=1.6cm]{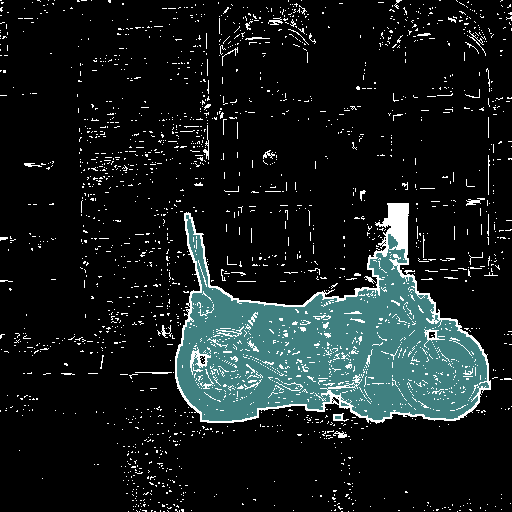}}
      {\includegraphics[width=1.6cm, height=1.6cm]{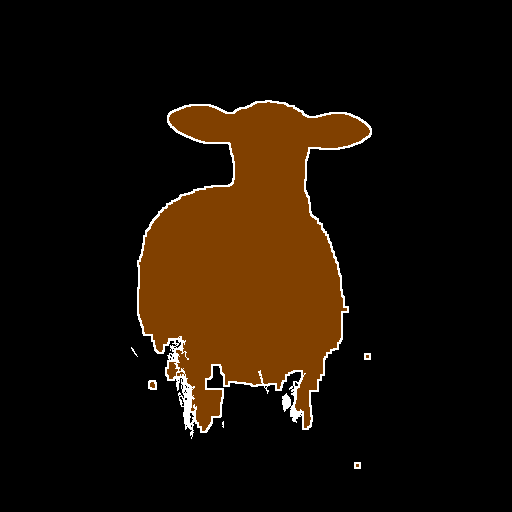}}
      {\includegraphics[width=1.6cm, height=1.6cm]{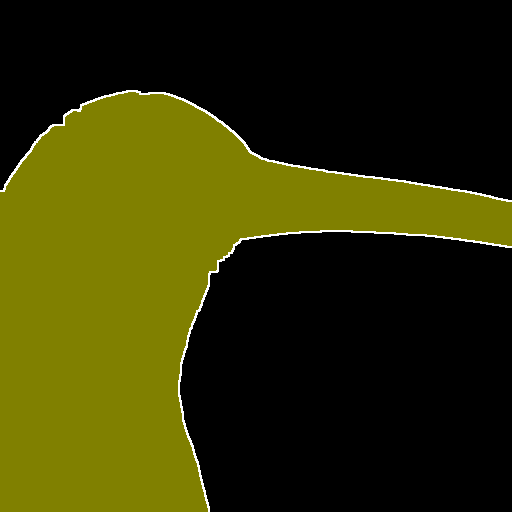}}
      {\includegraphics[width=1.6cm, height=1.6cm]{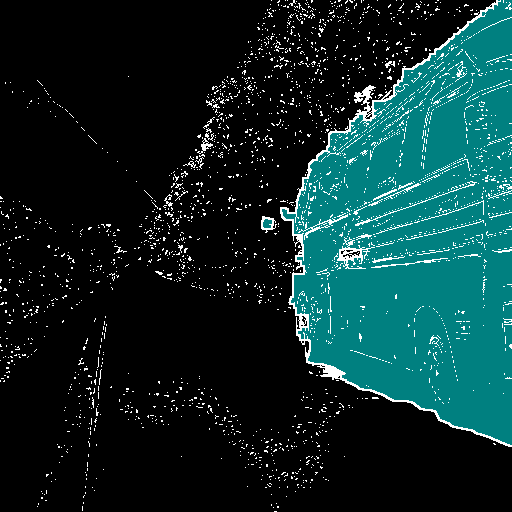}}
      {\includegraphics[width=1.6cm, height=1.6cm]{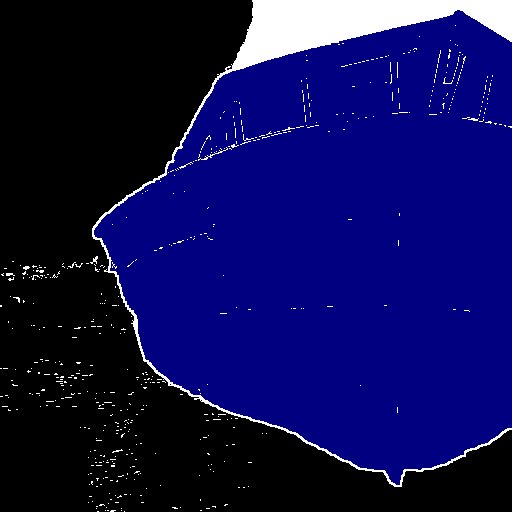}}
  \end{center}
      
  \caption{Examples from our Submitted Dataset}
  \label{fig:results}
  \end{figure}

\begin{table}[ht]
    \centering
    \begin{tabular}{l|c|c||l|c|c}
        \multicolumn{3}{c}{\textbf{Syntagen Private Test Set}} & \multicolumn{3}{c}{\textbf{Pascal VOC Validation Set}} \\ 
        \toprule
        \textbf{Class} & \textbf{IoU} & \textbf{Acc}  & \textbf{Class} & \textbf{IoU} & \textbf{Acc} \\ 
        \midrule
        background & 59.86 & 90.44  & background & 86.29 & 93.97 \\ 
        aeroplane  & 58.20 & 75.06  & aeroplane  & 77.37 & 92.89 \\ 
        bicycle    & 49.48 & 63.64  & bicycle    & 33.06 & 83.77 \\ 
        bird       & 49.11 & 50.66  & bird       & 75.64 & 83.44 \\ 
        boat       & 35.39 & 37.21  & boat       & 64.25 & 84.42 \\ 
        bottle     & 38.61 & 64.63  & bottle     & 65.46 & 87.60 \\ 
        bus        & 66.44 & 69.59  & bus        & 77.81 & 80.48 \\ 
        car        & 28.72 & 31.20  & car        & 69.94 & 76.28 \\ 
        cat        & 65.68 & 75.47  & cat        & 78.41 & 83.45 \\ 
        chair      & 14.69 & 17.43  & chair      & 26.84 & 48.69 \\ 
        cow        & 52.68 & 54.56  & cow        & 63.03 & 64.06 \\ 
        diningtable& 19.25 & 42.33  & diningtable& 41.72 & 74.39 \\ 
        dog        & 55.52 & 58.95  & dog        & 70.53 & 76.31 \\ 
        horse      & 63.67 & 74.73  & horse      & 67.72 & 81.12 \\ 
        motorbike  & 63.79 & 73.84  & motorbike  & 74.75 & 87.33 \\ 
        person     & 34.54 & 37.00  & person     & 42.94 & 50.95 \\ 
        pottedplant& 37.40 & 43.53  & pottedplant& 27.08 & 44.33 \\ 
        sheep      & 66.56 & 69.52  & sheep      & 67.30 & 74.88 \\ 
        sofa       & 11.60 & 12.07  & sofa       & 25.18 & 25.97 \\ 
        train      & 63.74 & 71.42  & train      & 69.58 & 84.25 \\ 
        tvmonitor  & 36.30 & 41.57  & tvmonitor  & 45.76 & 75.34 \\
        \midrule
        \textbf{mIoU} & \multicolumn{2}{c||}{46.25} & \textbf{mIoU} & \multicolumn{2}{c}{59.56} \\ 
        \bottomrule 
    \end{tabular} \\
      \caption{Comparison of Class-Wise Intersection over Union and Accuracy for the Private Syntagen Set (Left) and Pascal VOC Validation Set (Right)}
      \label{tab:comparison}

  \end{table}


\end{document}